\documentclass[onefignum,onetabnum]{siamonline190516}

\usepackage{lipsum}
\usepackage{amsfonts}
\usepackage{graphicx}
\usepackage{epstopdf}
\usepackage{algorithmic}
\ifpdf
  \DeclareGraphicsExtensions{.eps,.pdf,.png,.jpg}
\else
  \DeclareGraphicsExtensions{.eps}
\fi

\usepackage{enumitem}
\setlist[enumerate]{leftmargin=.5in}
\setlist[itemize]{leftmargin=.5in}

\newsiamremark{remark}{Remark}
\newsiamremark{hypothesis}{Hypothesis}
\crefname{hypothesis}{Hypothesis}{Hypotheses}
\newsiamthm{claim}{Claim}

\headers{Generalised learning in time-series}{M. S\"uzen, A. Yegenoglu}

\title{
Generalised learning of time-series: Ornstein-Uhlenbeck processes
}

\author{
Mehmet S{\"u}zen \thanks{\email{suzen@acm.org}}
\and Alper Yegenoglu 
}

\usepackage{amsopn}

\ifpdf
\hypersetup{
  pdftitle={Generalised learning of time-series: Ornstein-Uhlenbeck processes},
  pdfauthor={M. S{\u"zen, A. Yegenoglu}}
}
\fi

\begin{document}

\maketitle

\begin{abstract}
In machine learning, statistics, econometrics and statistical physics, cross-validation (CV) is used as a standard approach in quantifying the generalisation performance of a statistical model. A direct application of CV in time-series leads to the loss of serial correlations, a requirement of preserving any non-stationarity and the prediction of the past data using the future data. In this work, we propose a meta-algorithm called reconstructive cross validation (rCV ) that avoids all these issues. At first, k folds are formed with non-overlapping randomly selected subsets of the original time-series. Then, we generate k new partial time-series by removing data points from a given fold: every new partial time-series have missing points at random from a different entire fold. A suitable imputation or a smoothing technique is used to reconstruct k time-series. We call these reconstructions secondary models. Thereafter, we build the primary k time-series models using new time-series coming from the secondary models. The performance of the primary models are evaluated simultaneously by computing the deviations from the originally removed data points and out-of-sample (OSS) data.  Full cross-validation in time-series models can be practiced with rCV along with generating learning curves. 
\end{abstract}

\begin{keywords}
generalization, learning curves, time-series, cross-validation, Gaussian processes
\end{keywords}

\begin{AMS}
  37M10, 62M10, 62P35, 68T05, 68Q32
\end{AMS}

\section{Introduction}

The analyses of a temporal data generated by natural and technological systems are very
common (\cite{hamilton1994time}, \cite{richards2011a}, \cite{roberts2013a}). The primary
task of these analyses manifest as building time-series models for predicting future values.
In this context, evaluation of time-series model performances are not trivial,
specially in sparse settings (\cite{suzen16a}).
As a result, measuring the generalized performance of the time series model appears as an intricate problem (\cite{bergmeir2018}).

In general, cross-validation (CV) is used to judge generalisation ability of statistical
models (\cite{stone74}, \cite{efron83a}). It is also used to select the
best model (\cite{kohavi95}). However, usage of cross-validation for temporal data
is not directly practiced due to several issues that has to be addressed simultaneously:
serial correlations, stationarity properties and time ordering has to be retained.
These requirements create severe limitations on how CV should be applied for
time-series data. Instead, in practice usually a single out-of-sample (OOS) evaluation
is reported.
In the case of time-series CV, modified approaches have been proposed, such as shuffling chunks of time-series (\cite{politis94}) for stationary time-series, i.e. block bootstrap methods.
Other methods apply the CV directly on uncorrelated time-series (\cite{bergmeir2018}) or use a sliding OOS evaluation (\cite{hyndman18a}).

The learning curve is utilized when evaluating the relative performace of machine
learning algorithms (\cite{perlich2003}). Quantification of learning curves in
time-series are practiced rarely if at all.
Conventional learning curves are usually built via reducing the sample size of the dataset.
This is achieved by removing points randomly while reptively bulding the model.
By definition this accounts to supervised learning (\cite{mitchell97}).
Unfortunately, the aforementioned approach can not be used for building time-series learning curves directly, because it leads to similar issues when applying the CV directly.

We propose a meta-algorithm called {\it reconstructive Cross Validations} (rCV).
The method combines standard cross-validation (CV) and out-of-sample (OOS) evaluation
of the performance by introducing the reconstruction of missing points at random fold from CV,
i.e. imputation or smoothing.
This allows to generate $k$ new time-series and $k$ times OOS evaluations in achiving a time-series CV.
rCV does not require any assumption on the error or correlation structure of the time-series,
does not have any restriction on having stationary series and does not use
future data to predict past events in the main cross-validation procedure.
Moreover, time-series learning curves can be obtained by varying $k$
using $rCV$ repeatedly.

\subsection{Single model versus model selection}

The cross-validation procedure can be applied in two ways depending on how the data splits, i.e., $k$-folds, are used in the optimisation for model parameters:
Either the parameters are optimized $k$ times using different partitions of the data (\cite{kohavi95}) or $k$ different partitions are used within a single optimization (\cite{stone74}).
Common machine learning libraries implement $k$-fold cross-validation which produce $k$ different models.
Similarly rCV builds $k$ different
models in a supervised learning setting. Applying rCV to a variation of
cross-validation to produce a single model is also possible. This requires
a change on how underlying solvers interact with data. In this work, we
follow the mainstream practice.

\section{Proposed techniques}

Our contribution has two main implication in the generalised learning of time-series. On the one hand, we present
a generic procedure to do cross-validation for time-series models and on the other hand we suggest a technique to
build learning curves for time-series without need to reduce the sample size of the time-series data.

We concentrate on one dimensional time-series. 
Extensions to higher-dimensions should be self-evident. A series of numbers, $\mathbf{y} \in \mathbb{R}^{n}$, vector of length $n$,
$t_{i} \in \mathbb{R}^{n}$, where $y_{i} = y(t_{i})$ and $i \in \mathbb{Z}_{+}$,  $t_{n} > t_{n-1} > ... > t_{1}$
are considered. The ordered dataset reads as $(y_{i} , t_{i})$. This tuple of ordered numbers are regarded as a time-series,
$t$ is interpreted as the time evolution of $y_{i}$ and is usually expressed as $y_{i}(t_{i})$.

Out-of-sample (OOS)  data usually appears as a continuation of the past time-series, hence, a continuation of
$\mathbf{y}$ is defined as out-of-sample set, $\mathbf{w} \in \mathbb{R}^{p}$, vector of length $p$, $p > n$,
where $w_{j} = w(t_{j})$ and $j \in \mathbb{Z}_{+}$, where $j > p$ and $t_{m+p} > t_{p+n-1} > ... > t_{n+1}$.
The ordered OOS dataset reads as $(w_{j}, t_{j})$.

The construction of the cross-validated performance is measured and learning curves for time-series appear
as a meta-algorithm processing time-series $\mathbf{y}$ and $\mathbf{w}$.

\subsection{Reconstructive cross-validation for time-series}

The first step in rCV is to identify partitions of the time-series, here $\mathbf{y}$ as in the
conventional cross-validation (\cite{efron83a}). $k$ sets of partitions
of $\mathbf{y}$ are considered as $\mathbf{y}^{1}, \mathbf{y}^{2}, ...,\mathbf{y}^{k}$, each set have
randomly assigned $y_{i}$, and partitions are approximately equal in size
\begin{equation}
 |\mathbf{y}^{1}| \approx |\mathbf{y}^{2}| \approx ... \approx |\mathbf{y}^{k}|.
\end{equation}
A training-fold is defined as a union of all partitions except the removed partion,
\begin{equation}
Y^{l} = \bigcup_{l=1, l \ne m}^{k} \mathbf{y}^{l},
\end{equation}
The missing data appears on the corresponding removed partition $\mathbf{y}^{m}$.

Due to the ordered nature of the series, the standard CV approach is not used in
different folds which yields to an absurd situation of predicting past using future values.
To overcome this, a reconstruction of full training series $y$, denoted by $R^{j}$ can be introduced.
This can be thought of an imputation of missing data at random or smoothing in a Bayesian sense, using each
training-fold $Y^{j}$ via a secondary model $\mathbb{M}_{2}$. A technique could be interpolation or
more advanced filtering approaches like Kalman filtering resulting in
\begin{equation}
R^{m} = Y^{l} + \hat{\mathbf{y}}^{m}.
\end{equation}
The secondary model could retain the given points on the training-fold $Y^{j}$ in this approach.
$\hat{\mathbf{y}}^{m}$ is the reconstructed portion.

The total error due to reconstructed model $\mathbb{M}_{2}$ expressed as
$g_{r}(\mathbf{y}, \hat{\mathbf{y}_{}})$, here for example we write down as a
Mean Absolute Percent Error (MAPE) \cite{hyndman18a},
\begin{equation}
g_{r} = \frac{1}{k} \sum_{m=1}^{k}  |(\mathbf{y}^{m} - \hat{\mathbf{y}}^{m})|/\mathbf{y}^{m} .
\end{equation}
Different metrics can be used as well.
The primary model, $\mathbb{M}_{1}$, is build on each $R^{m}$ and predictions on the out-of-sample
set $\mathbf{w}$ are applied. This results in set of predictions $\hat{\mathbf{w}}^{m}$, the error
is expressed as $g_{p}(\mathbf{w}, \hat{\mathbf{w}_{}})$
\begin{equation}
g_{p} = \frac{1}{k} \sum_{q=1}^{k} (\mathbf{w} - \hat{\mathbf{w}}^{m}).
\end{equation}
The total error in rCV is computed as follows.
\begin{equation}
g_{rCV} = g_{r} \cdot g_{p}.
\end{equation}
The lower the number the better the generalisation, however both $g_{r}$ and $g_{p}$ should be judged seperately
to detect any anomalies. We have choosen $g_{rCV}$ as the multiplivative error of reconstruction and as the prediction errors,
so that it represents the weighted error. More complex schemes to estimate $g_{rCV}$ can be divised. Note that
both $g_{r}$ and $g_{p}$ are test errors in a conventional sense while both reconstruction and prediction
computations are performed on a Gaussian process that parameters are fixed, i.e., corresponding to
Ornstein-Uhlenbeck processes.

\subsection{Learning Curves for time-series}

Time-series learning curves are not that common \cite{giola21}
This is  due to fact that data sample sizes are limited and and the conventional procedure introduces the same issues as mentioned earlier when the direct CV is applied on the time-series datasets.
However with the $rCV$ approach one can build
a learning curve $L$ based on different number of folds
\begin{equation}
L^{err}(k) = g_{err}^{k}.
\end{equation}
$err$ corresponds to the error measure used with different errors over a range of different $k$ values. The error
terms can be any of the errors defined above $g_{rCV}$, $g_{r}$ or $g_{p}$. Unlike conventional
learning curves build upon reducing sample-size (\cite{perlich2003}), $L^{err}(k)$ is constructed while
retaining the sample-size of the original time-series $\mathbf{y}$. The reason behind this fixed
sample size is that the number of missing data at random on the reconstructed folds will decrease with an increasing number of the folds.
Hence, the change in the missing sample size along different $k$ values mimics the experience level in the learning curves.
A combination of thee reconstruction and prediction errors as performance measure will be affected by the by changing the number of the folds.
As a consequence the learning curve known from the basic definition of supervised learning (\cite{mitchell97}) is achived.

\begin{figure}
  \includegraphics[width=0.45\textwidth]{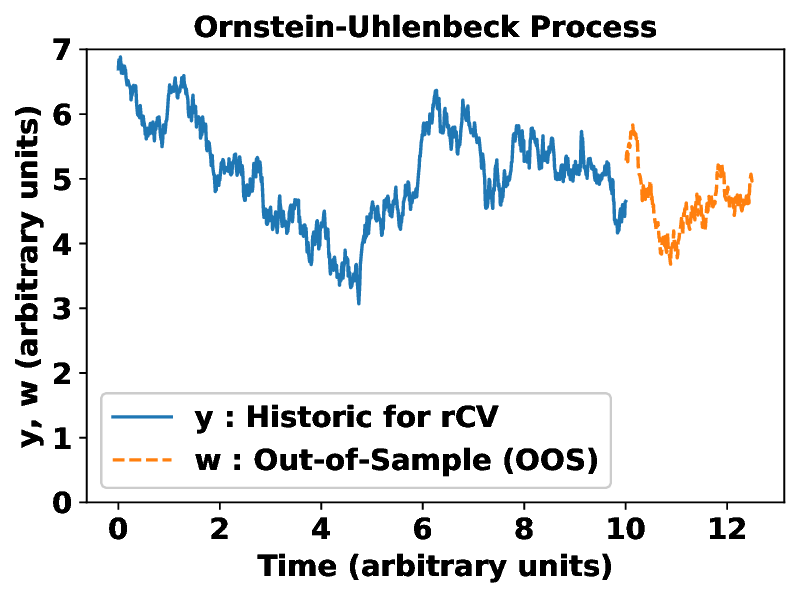}
  \includegraphics[width=0.45\textwidth]{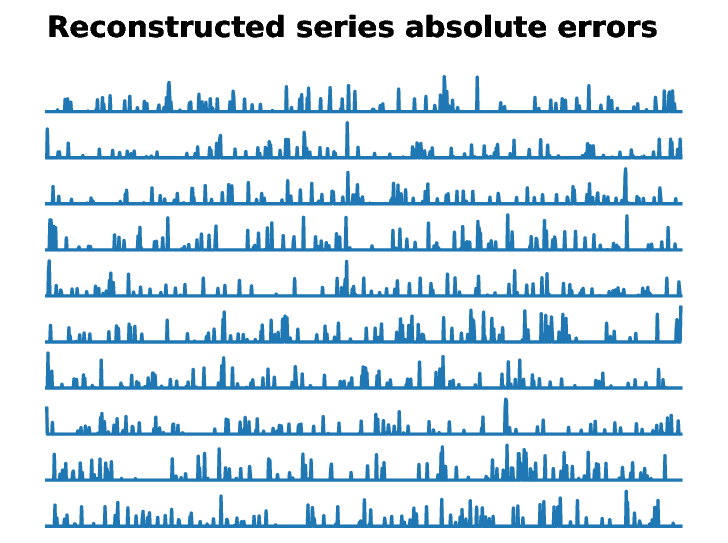}
  \caption{Simulated Ornstein-Uhlenbeck data corresponding to $\mathbf{y}$ and $\mathbf{w}$ series in our formulation (left).
           10-fold reconstruction absolute errors, difference between reference and imputed time series $R^{m}$, mean
           difference is $0.014$ (right).}
\end{figure}

\section{Experimental Setup}

We have demonstrated the utility of our technique using a specific kernel in a Gaussian process setting (\cite{williams2006a}).
This corresponds to an {\it Ornstein-Uhlenbeck process}  and is used in the description of Brownian motion in statistical physics (\cite{gardiner09}). The learning task is aiming to predict the OOS data $\mathbf{w}$ using past series $\mathbf{y}$.
Note that, for this particular setup the kernel parameters are already fixed, but this is not a
restriction on $rCV$. Any appropriate learning tasks can be put in the $rCV$ framework.

\subsection{Ornstein-Uhlenbeck process}

One can generate an {\it Ornstein-Uhlenbeck process} drawing numbers from multivariate Gaussian with a specific covariance structure,
\begin{equation}
\mathbf{y}_{ou}(t_{i}) \sim \mathcal{N}(\mathbf{\mu}, \Sigma).
\end{equation}
All $\mathbf{\mu}$ are set at $5.0$, and $\Sigma$ is built using the kernel $\exp(-\mathbf{D}^{ii}/2.0)$ where $\mathbf{D}^{ii}$ is the distance matrix constructed over time-points. $\mathbf{D}^{ii}$ is a symmetric matrix,
$$
\mathbf{D}^{ii} =
\left( \begin{matrix}
    0             & |t_{1}-t_{2}| & |t_{1}-t_{3}| & \dots  & |t_{1}-t_{n}| \cr
    |t_{2}-t_{1}| & 0             & |t_{2}-t_{3}| & \dots  & |t_{2}-t_{n}| \cr
    \vdots        & \vdots        & \vdots        & \ddots & \vdots \cr
    |t_{n}-t_{1}| & |t_{n}-t_{2}| & |t_{n}-t_{3}| & \dots  & 0 
    \end{matrix}
\right)$$
We generated an Ornstein-Uhlenbeck (OU) time-series for $1000$ regular time points with a spacing of $\Delta t = 0.1$, with different length scales, mean values $\mu$, and additional $250$ time points for the prediction task (c.f. Figure 1).

\subsection{Reconstructive cross-validation}

We apply our meta-algorithm to construct both primary and secondory models using Gaussian process predictions with unit regularisation, which is formulated as follows: Given $(y_{i}, t_{i})$ ordered pairs as time-series, we aim at inferring, i.e., reconstructing, missing values at time-points $t_{j}$. The missing values $y_{j}$ can be identified via a Bayesian interpretation of Kernel regularisation,
\begin{equation}
y_{j} = K_{ji} L^{-1} y_{i}
\end{equation}
where $L=K_{ii} + \mathbb{I}$. The kernel matrices $K_{ji}$ and  $K_{ii}$ are build via the kernel $\exp(-\mathbf{D}^{ji}/2.0)$
where $\mathbf{D}^{ji}$ is the distance matrix over time-points of missing at random folds and the other folds and
$\exp(-\mathbf{D}^{ii}/2.0)$ respectively. A secondary model is used to reconstruct $R^{m}$. The absolute errors in the 10-folds are shown in Figure 1.

A similar procedure is followed in predicting the OOS vector $\mathbf{w}$. The reconstruction error, prediction error and rCV error were computed as 0.029, 0.468 and 0.013 respectively. Note that rCV error is not a MAPE but a measure of generalisation. The high prediction error is attributed to a long time horizon we choose. In practice, a much shorter time-horizon should be used.

\begin{figure}
  \includegraphics[width=0.45\textwidth]{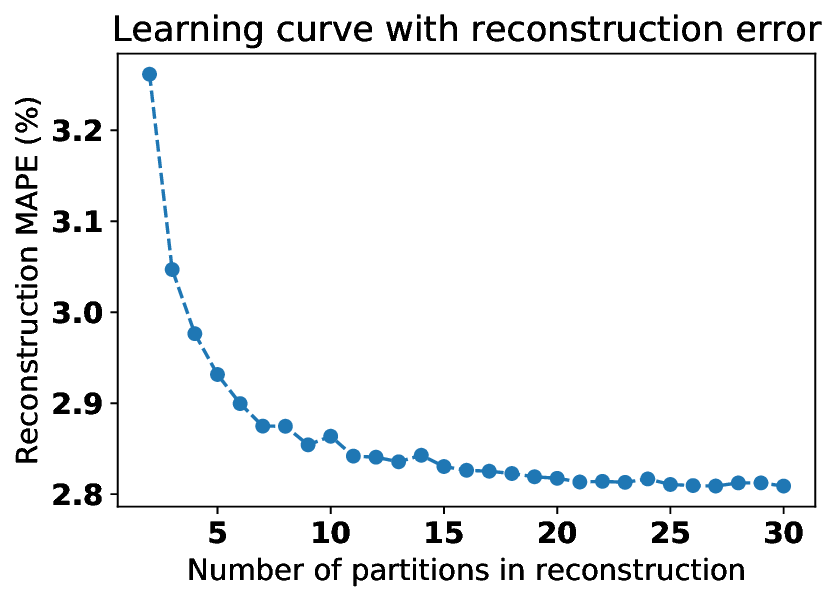}
  \includegraphics[width=0.45\textwidth]{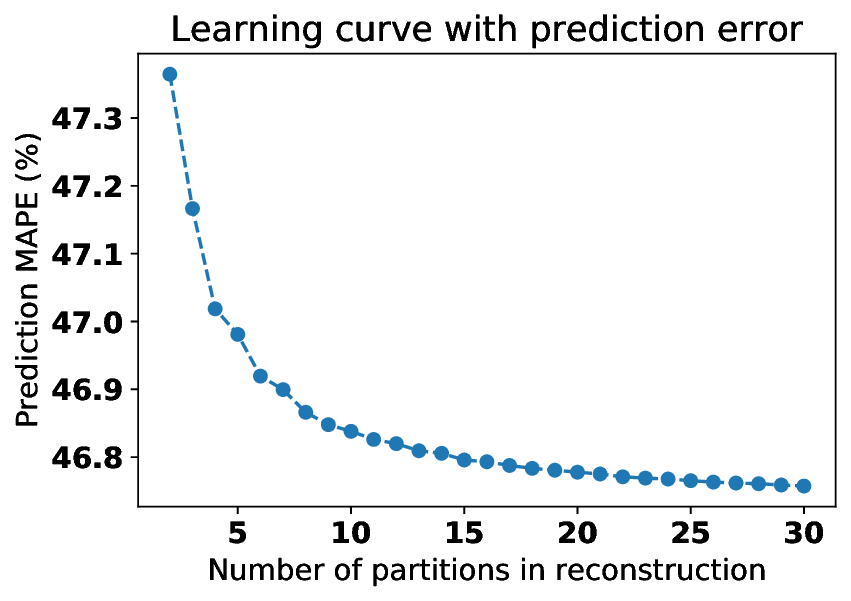}
  \caption{
           Different test learning curves in our meta-algorithm. Based on the reconstruction error (left) and the prediction error (right).
           Increasing number of folds indicates less number of points to reconstruct, i.e., larger sample-size in a traditional sense.
          }
\end{figure}

\section{Conclusion}

We have presented a framework with a canonical process from Physics, the Ornstein-Uhlenbeck process, that helps us
to perform generalised learning in time-series without any restriction on the stationarity and to retain the
serial correlations order in the original dataset.
The approach entails applying cross-validation in a modified fashion
in combination with OOS estimate of the performance and
reconstructing missing random fold instances by applying a secondary model. rCV, also allows to generate a learning curve for time-series.

The meta-algorithm we developed in this work can be used with any other sequence learning algorithms, such as LSTM.
We have choosen the Gaussian processs for the reconstruction and prediction tasks because of its minimal requirements to implement and demonstrate it within our framework.
Further implementation of the meta-algorithm in a generic setting is possible without embedding the learning algorithm into the rCV procedure.

\bibliographystyle{siamplain}
\bibliography{rcv}

\end{document}